\documentclass[conference]{IEEEtran}
\usepackage[noadjust]{cite}
\usepackage{amsmath}
\usepackage{array}
\usepackage{longtable}
\usepackage{amsmath}
\usepackage{epsfig,subfigure}
\usepackage{graphicx}

\usepackage{cite}
\usepackage{array}
\usepackage{algorithmic}
\usepackage{algorithm}
\usepackage{cite}
\usepackage{longtable}
\usepackage{setspace}

\usepackage{amsmath}

\begin{document}
\title{A Lower Bound for the Variance of  Estimators for  Nakagami-$m$ Distribution}
\author   {\IEEEauthorblockN{Rangeet Mitra}\IEEEauthorblockA{Amity University, Gurgaon\\India\\Email: mitra.rangeet@gmail.com} \and \IEEEauthorblockN{Amit Kumar Mishra}\IEEEauthorblockA{University of Cape Town\\Rhondebosch, South Africa\\Email: amit.india@gmail.com}  \and \IEEEauthorblockN{Tarun Choubisa}\IEEEauthorblockA{IISC Bangalore\\Email:choubisa.tarun@gmail.com}
\thanks{Rangeet Mitra, Tarun Choubisa and Amit Kumar Mishra are  with
Electronics  and  Communiwcation  Engg. Department,
Indian Institute  of Technology,  Guwahati,  India.
E-mail: rangeet@iitg.ernet.in, c.tarun@iitg.ernet.in and
amishra@iitg.ernet.in.}}
\maketitle
\begin{abstract}
Recently we have proposed a maximum-likelihood iterative algorithm for estimation of parameters of the Nakagami-$m$ distribution. This  technique performs better than state of art estimation techniques for this distribution.
This could be of particular use in low-data/block based estimation problems.
In these scenarios, the estimator should be able to give accurate estimates (in the mean square sense) with less amount of data. Also, the estimates should improve with increase in number of blocks received.
In this paper, we see through our simulations, that our proposal is well designed for meeting such requirements.
Further, it is well known in the literature that an efficient estimator does not exist for the Nakagami-$m$ distribution. 
In this paper, we also derive a theoretical expression for the variance of our proposed estimator. We find that this expression clearly fits the experimental curve for the variance of the proposed estimator. This expression is pretty close to the Cramer Rao Lower Bound (CRLB).
\end{abstract}
\begin{IEEEkeywords}
 Nakagami-$m$ distribution, Maximum-Likelihood, Cramer-Rao Lower Bound.
\end{IEEEkeywords}
\section{Introduction}

 Nakagami-$m$  distribution is a versatile distribution which is used for modelling a variety of physical phenomena in communication engineering.
 An algorithm for estimation of the parameters of  Nakagami-$m$ distribution has been recently proposed in \cite{mitra_mishra}.
 However many seminal papers like \cite{Cheng2001} have adressed this topic as well.
  The types of estimators proposed in the literature can be catagorised into  two types: a)moment-based \cite{Cheng2002,Chen2005,Ko2003,Chen2005a,Had2007,Wang2012} b) algebraic maximal likelihood \cite{Cheng2001,Zhang2002} methods.
  In \cite{Zhang2002}  the authors have studied many pre-existing parameter estimators for the Nakagami-$m$ distribution which belonged to both of the above mentioned category of estimators.
The problem with moment based estimators is that it requires a large number of samples to estimate the moments accurately.
On the other hand, the algebraic methods suffer from errors due to truncation of Taylor's series.

An interesting work, \cite{Zhang2002}, concludes that none of the many estimators discussed in it gives better performance than the Greenwood and Durand estimator \cite{Zhang2002} (which will be one of the many algorithms against which we will compare our approach in this paper).
Also, \cite{Chen2005,Chen2005a} deal with parameter estimation of noisy Nakagami-$m$ signals.
However, in noisy conditions, a) the distribution is no longer Nakagami-$m$, hence it is out of context for our study for now; and b) It uses moment based parameter estimates which require a significant number of samples to achieve a low variance; and lastly c) it needs knowledge of the variance of noise which corrupts the signal which could be Gaussian/Non-Gaussian.

The suitability of our approach in \cite{mitra_mishra} against these previously existing algorithms has not been tested.
In the current paper we we derive a new simple theoretical lower bound for variance of the $m$ parameter estimates for this algorithm.
It is seen that our proposed algorithm has two advantages; viz. a) It performs better than other estimators with low amounts of data in highly testing scenarios; and, b) as it is an online algorithm which considers one sample at a time the parameter $\Delta$ in \cite{Cheng2001} is zero.
Also, it does not make any algebraic approximations. This prevents error propagation between blocks in block processing scenarios. Due to these intuitive reasons, our proposed algorithm is better than the state of art in low-data/block scenarios.


The rest of the paper is organised as follows.
Section II describes the conditions in which we simulated our algorithm. Section III describes the final algorithm. Section IV gives some heuristics which should be used. Section V gives the comparison of the variance of our algorithm with some existing estimators. Finally in the Appendix we give an expression for the Cramer-Rao lower bound (CRLB) based lower bound of the variance of our algorithm.

\section{Nakagami-$m$ Distribution and the Porposed Parameter Estimation Algorithm}
The multivariate Nakagami-$m$ distriution is approximated as the product of $L$ individual Nakagami distributions (assuming that the data is i.i.d).
\begin{equation}
P(\textbf{x})=\Pi_{i=1}^{L}\frac{2^L}{\Gamma(m_{i})\sigma_{i}^{m_{i}}}
x_{i}^{2 m_{i}-1} \exp(-\frac{x_{i}^{2}}{\sigma_i} )
\end{equation}

Here, $\sigma_{i}$ and $m_{i}$ are the spread factors and the centrality factors for each of the component Nakagami-$m$ distributions.

As we take the derivative of the log-likelihood with respect to each of the components, 
and equate it to zero and solve the differential equation for the $\Gamma(m)$, we get the following equations, which are solved numerically.
\begin{equation}
\Gamma(m_{i})=\frac{x_{i}^{2 m_{i}}}{\sigma_{i}^{m_{i}}}
\end{equation}
and,
\begin{equation}
\sigma_{i}=\frac{1}{m_{i}}x_{i}^2
\end{equation}
 Details of the derivation are provided in \cite{mitra_mishra}.

\section{Simulation Conditions}\label{algo}
In this section, we give a description of our simulation set up for comparison of our algorithm with the state of art.
At a time we receive a limited parallel block of data. The deployed (``competing") algorithms  have to infer from a subset of such limited data blocks. A desirable property of an estimation algorithm in such a case is increased learnability with the number of blocks it sees (apart from less Mean Squared Error (MSE)). Further, in such limited data conditions, we may like to use a moving average estimator (which is the most simple and recursive approach to smoothing) to smooth the estimates. However, if the deployed algorithm has errors from factors like truncation of Taylor's series, these errors will get propagated from block to block.

This intuition is caught by our algorithm. Our algorithm never uses series truncation. It relies on numerical methods and hence has the ability to learn improvably from multiple blocks. Also, our algorithm does not require large datasets  for the moments to converge (unlike its moment based counterparts). Our simulations presented in this section confirm to this intuitive reasons for the success of our algorithm against its counterparts.

The algorithm to be followed in our comparison of parameter estimation methods can be given as below.

\begin{algorithm}                      
\caption{Parameter Estimation for  $m$ for $\{\textbf{x}\}_{i=1}^{N}$}          
\label{basic}                           
\begin{algorithmic}[1]
\FOR{$i = 1$ to $N$}
\IF{$i==1$}
\STATE Estimate $\hat{m}_{1}$ via any of the candidate estimators described.
\ELSE
\STATE Estimate $\hat{m}_{i}$ via any of the candidate estimators.
\STATE Update current mean as $\hat{m}_{i} = \frac{i-1}{i}\hat{m}_{i-1}+\frac{1}{i}\hat{m}_{i}$
\ENDIF
\ENDFOR
\end{algorithmic}
\end{algorithm}


\subsection{Heuristics Considered}
As the iterative equations given in \cite{mitra_mishra} are non-convex, they may be susceptible to local minima. Hence we must run the algorithms a number of times and take centrality measures(mean/median/mode) of the obtained values.

Also, to help all the estimators learn better (and also facilitate comparison on equal terms) the estimated values obtained by all the techniques considered in the next section are averaged over the blocks by the well known recursive-mean filter (i.e. the sample mean calculated recursively).

%
%

\section{Comparison of all the Estimators}\label{est_disc}
\begin{figure}
  \centering
  \includegraphics[width=10cm, height = 10cm]{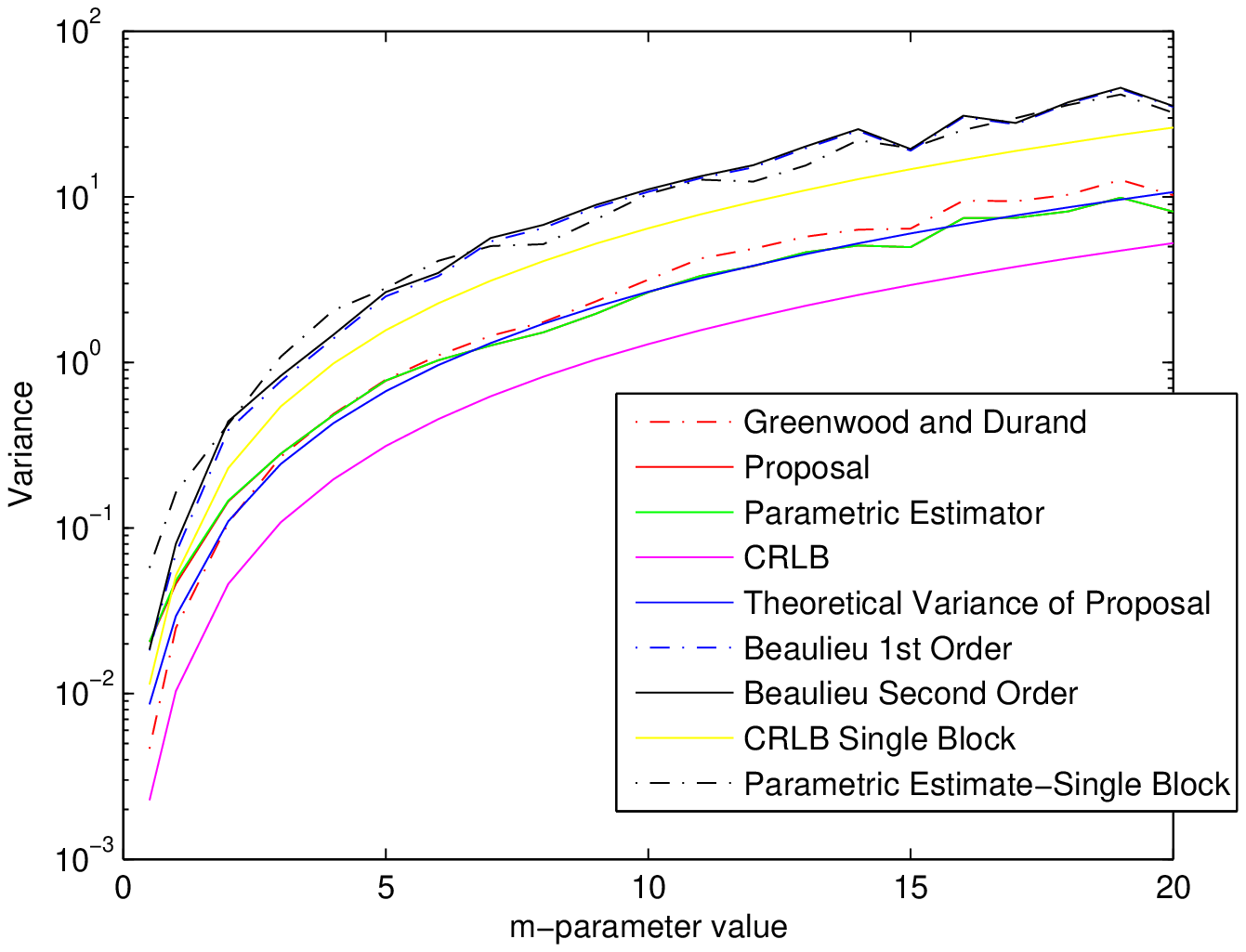}\\
  \caption{Comparison of various algorithms with respect to the proposal(Block Window Size 30X5)}\label{comp}
\end{figure}

We can see from figure \ref{comp} that our estimator learns with number of blocks it sees, i.e. its performance lies between the CRLB of the single block and that of the whole block. Also its performance is better than the estimators proposed in \cite{Cheng2001} (which were based on first order and second order Taylor approximations). Also its performance is better than the Greenwood and Durand estimators given in \cite{Zhang2002}(which were the most superior in that paper) and equivalent to the moment based estimator in most of the $m$ regimes. Hence our algorithm has sufficient credibility in scenarios where block data processing is common like Orthogonal Frequency Division multiplexing (OFDM) and data is limited.

\begin{figure}
  \centering
  \includegraphics[width=8cm, height = 8cm]{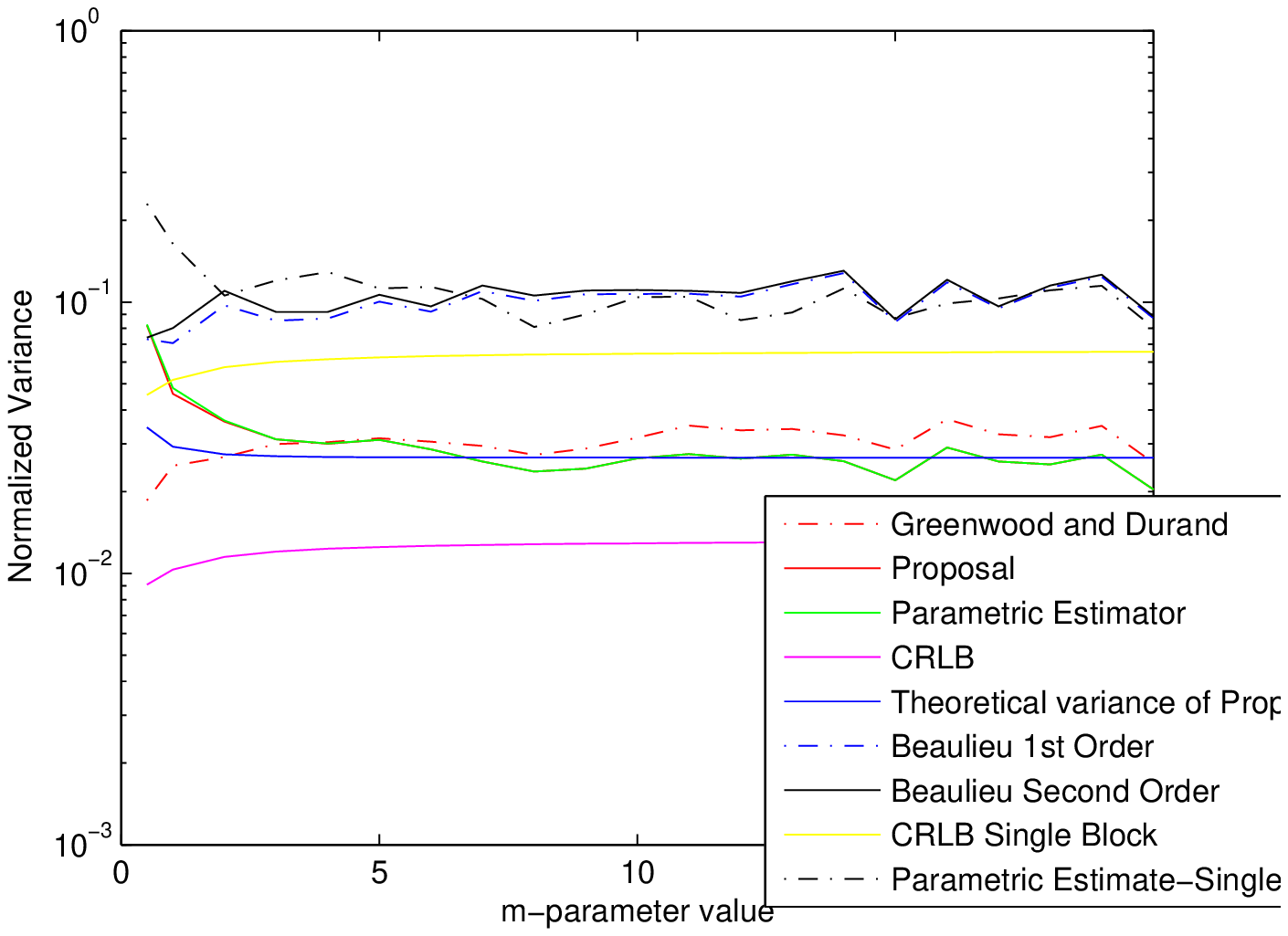}\\
  \caption{Comparison of various algorithms with respect to the proposal(Block Window Size 30X5)}\label{comp1}
\end{figure}

\begin{figure}
  \centering
  \includegraphics[width=8cm, height = 8cm]{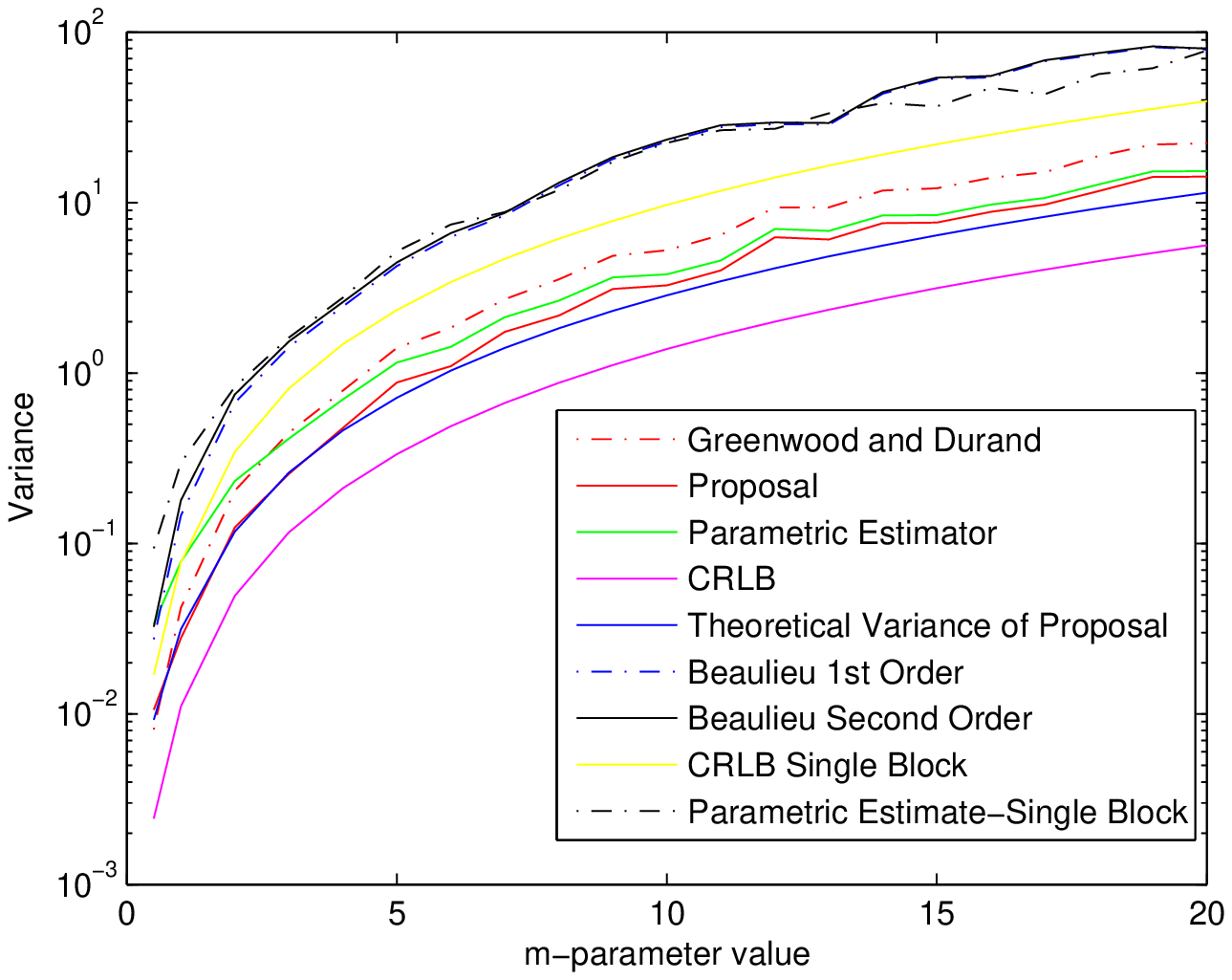}\\
  \caption{Comparison of various algorithms with respect to the proposal(Block Window Size 20X7)}\label{comp2}
\end{figure}

\begin{figure}
  \centering
  \includegraphics[width=8cm, height = 8cm]{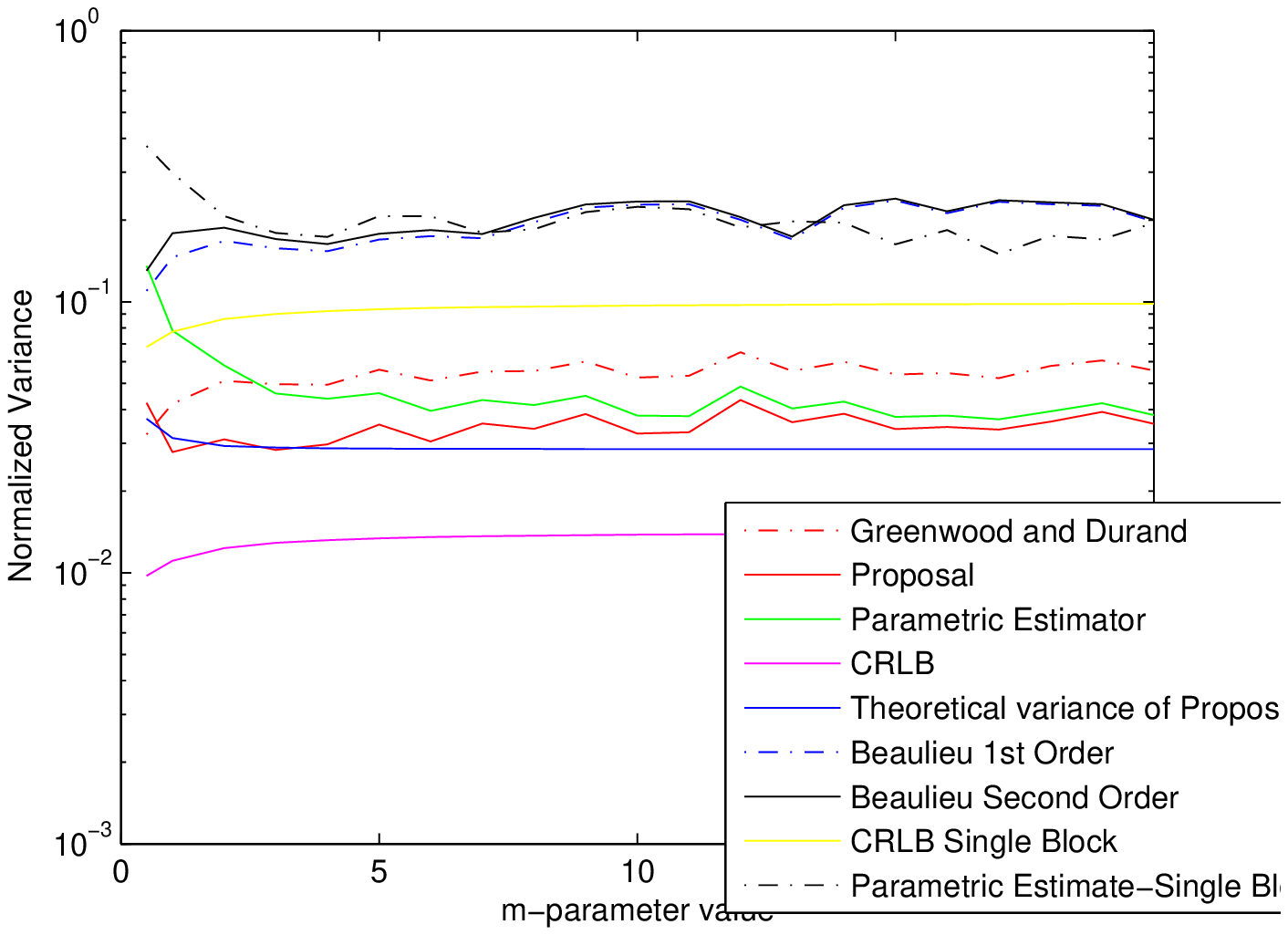}\\
  \caption{Comparison of various algorithms with respect to the proposal(Block Window Size 20X7)}\label{comp3}
\end{figure}

For the sake of completeness, the normalized variance plots are also shown in Fig. \ref{comp1}. In Figs. \ref{comp2} and \ref{comp3} we compare the performance of all the estimators in the setting $20X7$ block size (i.e in reference of Sec. \ref{algo}, the $i^{th}$ block is of size $20$ and $N=7$). We see that in this scenario, our estimator even outperforms the moment based estimator.

\section{A Possible Practical Application}
Consider the case of Orthogonal Frequency Division Multiplexing(OFDM). After FFT at the receiver we receive a block of data which is converted to serial form via parallel to serial(P/S) converter. When the number of points obtained in each conversion is less, the variance of state of art estimators increases (apart from the pre-mentioned errors stemming from algebraic approximations as in \cite{Cheng2001}).  In such cases, we must use an algorithm which is a) exact b) iterative (for robustness and tracking non-stationarity). Also, one of the secondary objectives of our algorithm is to omit the role of P/S converter before parameter estimation for performing detection.

\section{Conclusion}

In this paper we have found that our proposed algorithm has better performance than some of the popular estimators. We give a theoretical expression for the variance of our estimator in the appendix \ref{derivation}. We see that the variance of our estimates obtained via simulations closely matches the theoretical expression.

\section{Review of Hidden Markov Random Fields(HMRF) as Applied to Image Segmentation}
Basically in the project in \cite{DBLP:journals/corr/abs-1207-3510}, the image segmentation algorithm has the following assumptions:
Let $\textbf{y}$ be the set of image pixels and $\textbf{x}$ be the set of labels which are initialized similarly by k-means algorithm and are later inferred again by (Expectation-Maximization) EM algorithm. Then for a particular set of similarly initialized $\textbf{x}$,

\begin{equation}
p(y \mid x) = G(y,\mu,\sigma)
\end{equation}
where $G$ is a Gaussian with mean $\mu$ and variance $\sigma$.

Under the i.i.d assumption,
\begin{equation}\label{cnd}
p(y \mid x) = \prod_{i}G(y_{i},\mu_{i},\sigma_{i}).
\end{equation}
Clique (denoted by $c$, which is a fully connected subset of a graph) potential on a 4-neighborhood of pixels is defined as follows:
\begin{equation}\label{pri}
V_{c}(x_{a},x_{b}) = 1 - \delta(x_{a}-x_{b}).
\end{equation}
Here the $\delta$ denotes a sample impulse to enforce a constraint/prior of nearby pixels being assigned the same label with higher probability.
These clique potentials are then added to the potentials of the energy in Eqn. \ref{cnd} used to develop priors. The total energy is the sum of the following two energies,
\begin{equation}
U = log(p(y \mid x)) + log(\Sigma_{c}V_{c})
\end{equation}

The labels are then found by the step,
\begin{equation}
x^{*} = argmin_{x} U
\end{equation}

From the new inferred labels new priors are inferred:
\begin{equation}\label{pdf}
p^{'}(y) = \frac{exp(\Sigma_{c}V_{c}(x_{a}^{*},x_{b}^{*}))}{Z}
\end{equation}

$x_{a}$ and $x_{b}$ are two neighboring pixels connected by a Markov graph as in \cite{DBLP:journals/corr/abs-1207-3510} forming a clique.
$Z$ is the summation of the numerator of Equation \ref{pdf} over all possible cliques. It is basically a normalization constant to make $p^{'}(y)$
a valid pdf.
Using these the sample mean and variance are inferred by the following equations:

\begin{equation}
\mu^{*} = \Sigma_{y}p^{'}(y)y
\end{equation}

\begin{equation}
\sigma^{*} = \Sigma_{y}p^{'}(y)(y - \mu^{*})^{2}
\end{equation}
\section{Modified Nakagami-$m$ based MRF segmentation}
In our modification, we allow the conditional pdf to be Nakagami-$m$ distributed, i.e.,
\begin{equation}\label{cnd}
g(y|x)=\Pi_{i=1}^{L}\frac{2^L}{\Gamma(m_{i})\sigma_{i}^{m_{i}}}
y_{i}^{2 m_{i}-1} \exp(-\frac{y_{i}^{2}}{\sigma_i} )
\end{equation}

Given a particular labeling pattern \textbf{x}, the parameters are inferred from the algorithm in Sec. \ref{algo}(by our previous simulations in this paper our algorithm comes closest to the CRLB than any other compared estimator in limited data) for each of the labels.

The labels, in turn are again inferred from the minimization of the following potential,

\begin{equation}
U = log(g(y \mid x)) + log(\Sigma_{c}V_{c})
\end{equation}

\section{Segmentation Results}
\begin{figure}[h]

\centering

\subfigure[Performance of Gaussian likelihood based Segmentation]{

    \includegraphics[width=6cm, height = 6cm]{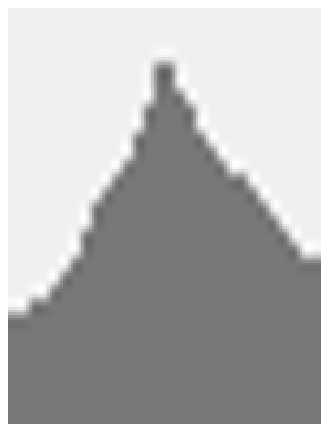}\quad
    \label{fig:subfig1}

}

\subfigure[Original Image]{

\includegraphics[width=6cm, height = 6cm]{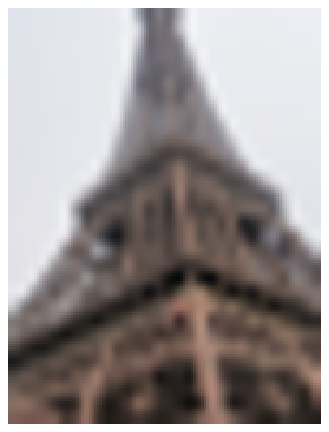}\quad
    \label{fig:subfig2}

}

\subfigure[Performance of Nakagami-m likelihood based Segmentation]{

\includegraphics[width=6cm, height = 6cm]{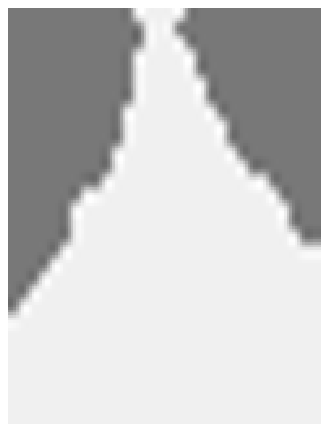}
    \label{fig:subfig3}

}
\end{figure}

\begin{figure}[h]

\centering

\subfigure[Performance of Gaussian likelihood based Segmentation]{

    \includegraphics[width=6cm, height = 6cm]{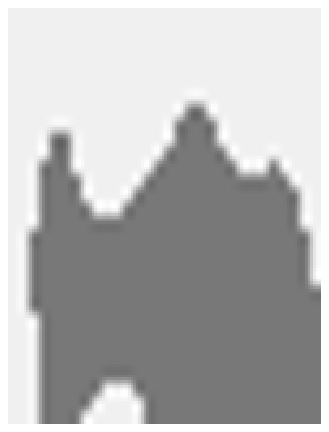}
    \label{fig:subfig4}

}

\subfigure[Original Image]{

\includegraphics[width=6cm, height = 6cm]{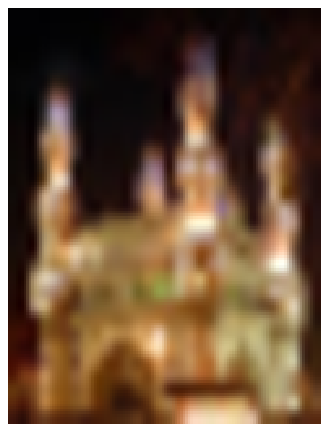}
    \label{fig:subfig5}

}

\subfigure[Performance of Nakagami-m likelihood based Segmentation]{

\includegraphics[width=6cm, height = 6cm]{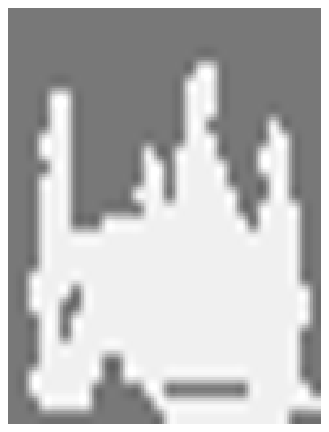}
    \label{fig:subfig6}

}

\caption[Image Segmentation Performance Comparison]

\label{fig:subfigureExample}

\end{figure}

A $600X338$ RGB image was taken and was deliberately scaled down to $30X30$ size to see performance comparisons in limited data. Then it was blurred by a 3X3 Gaussian blur. After segmentation, the image is resized to $160X120$ for ease of viewing. The performance of the Gaussian based HMRF and Nakagami-$m$ HMRF are compared. We can see from Fig. \ref{fig:subfig1} that the top of the tower is chopped off in the process of segmentation. Fig. \ref{fig:subfig2} shows the original (resized) image. However the entire tower is retained in \ref{fig:subfig3} even after nakagami-$m$ segmentation.   We can also see better segementation result in case of ``Charminar" image from Figs. \ref{fig:subfig4},\ref{fig:subfig5},\ref{fig:subfig6}.

\appendix \label{derivation}
\section{On the CRLB}

We know that $log(\Gamma(m))$ is a log-convex function. This implies,

\begin{equation}
log(\Gamma(m+0.5)) \geq log(\Gamma(m)) + 0.5 \psi(m)
\end{equation}

or,
\begin{equation}
2(log(\Gamma(m+0.5))-log(\Gamma(m))) \geq \psi(m)
\end{equation}
or,

\begin{equation}
\psi(m) \leq 2log(\frac{\Gamma(m+0.5)\sigma^{0.5}}{\Gamma(m)}) - log(\sigma)
\end{equation}

or,

\begin{equation}\label{jensen}
\psi(m) \leq 2log(E[x]) - log(\sigma)
\end{equation}

By Jensen's inequality,
\begin{equation}
E[log(x)] \leq log(E[x])
\end{equation}

This implies that the inequality in Equation. \label{jensen} must hold for the least possible value of $log(E[x])$. Thus,

\begin{equation}
\psi(m) \leq log(\frac{x^2}{\sigma})
\end{equation}

or $\psi(m)$ lies in a $\delta_{1}$ neighborhood of $log(\frac{x^2}{\sigma})$, i.e.,

\begin{equation}
\psi(m) + \delta_{1} = E[log(\frac{x^2}{\sigma})]
\end{equation}

where $\delta_{1}$ is a non-zero number.

From concavity of $\psi(m)$,

\begin{equation}
\psi^{'}(m) \geq 2(\psi(m+0.5)-\psi(m))
\end{equation}

or,

\begin{equation}
\psi^{'}(m) = 2(\psi(m+0.5)-\psi(m)) + \delta_{2}
\end{equation}

It is trivial to see,

\begin{equation}
\delta_{1}\rightarrow 0 \Leftrightarrow \delta_{2}\rightarrow 0
\end{equation}

because,

\begin{equation}
\psi^{'}(m) = 2(\psi(m+0.5)-\psi(m)) \Leftrightarrow  \psi(m) \approx log(\frac{x^2}{\sigma})
\end{equation}

Then, after a simple analysis our approximate lower bound for variance becomes,
\begin{equation}
CRLB^{'} = \frac{1}{N[2\psi(m+0.5)-2\psi(m)-\frac{1}{m}]}
\end{equation}

This is the modified expression for the lower bound for the variance of this estimator.

This expression is obtained by putting the minimum possible value of the $\psi^{'}(m)$. This situation occurs iff the equations in \cite{mitra_mishra} are exactly solved. This expression is clearly greater than the CRLB for the Nakagami-$m$ distribution.

We can see from the previous simulations that our estimator almost faithfully follows this curve as a function of $m$.

\bibliographystyle{ieeetran}
\bibliography{ref1}

\begin{thebibliography}{10}
\providecommand{\url}[1]{#1}
\csname url@samestyle\endcsname
\providecommand{\newblock}{\relax}
\providecommand{\bibinfo}[2]{#2}
\providecommand{\BIBentrySTDinterwordspacing}{\spaceskip=0pt\relax}
\providecommand{\BIBentryALTinterwordstretchfactor}{4}
\providecommand{\BIBentryALTinterwordspacing}{\spaceskip=\fontdimen2\font plus
\BIBentryALTinterwordstretchfactor\fontdimen3\font minus
  \fontdimen4\font\relax}
\providecommand{\BIBforeignlanguage}[2]{{%
\expandafter\ifx\csname l@#1\endcsname\relax
\typeout{** WARNING: IEEEtran.bst: No hyphenation pattern has been}%
\typeout{** loaded for the language `#1'. Using the pattern for}%
\typeout{** the default language instead.}%
\else
\language=\csname l@#1\endcsname
\fi
#2}}
\providecommand{\BIBdecl}{\relax}
\BIBdecl

\bibitem{mitra_mishra}
R.~Mitra, A.~Mishra, and T.~Choubisa, ``Maximum likelihood estimate of
  parameters of nakagami-m distribution,'' in \emph{Communications, Devices and
  Intelligent Systems (CODIS), 2012 International Conference on}, 2012, pp.
  9--12.

\bibitem{Cheng2001}
J.~Cheng, S.~Member, and N.~C. Beaulieu, ``{Maximum-Likelihood Based Estimation
  of the Nakagami m Parameter},'' vol.~5, no.~3, pp. 101--103, 2001.

\bibitem{Cheng2002}
\BIBentryALTinterwordspacing
J.~Cheng and N.~Beaulieu, ``{Generalized moment estimators for the Nakagami
  fading parameter},'' \emph{IEEE Communications Letters}, vol.~6, no.~4, pp.
  144--146, Apr. 2002. [Online]. Available:
  \url{http://ieeexplore.ieee.org/lpdocs/epic03/wrapper.htm?arnumber=996038}
\BIBentrySTDinterwordspacing

\bibitem{Chen2005}
Y.~Chen, S.~Member, and N.~C. Beaulieu, ``{Novel Nakagami- m Parameter
  Estimator for Noisy Channel Samples},'' vol.~9, no.~5, pp. 417--419, 2005.

\bibitem{Ko2003}
Y.-C. Ko and M.-S. Alouini, ``{Estimation of Nakagami-m Fading Channel
  Parameters With Application to Optimized Transmitter Diversity Systems},''
  vol.~2, no.~2, pp. 250--259, 2003.

\bibitem{Chen2005a}
Y.~Chen, S.~Member, and N.~C. Beaulieu, ``{Estimators Using Noisy Channel
  Samples for Fading Distribution Parameters},'' vol.~53, no.~8, pp.
  1274--1277, 2005.

\bibitem{Had2007}
H.~Nasuf \emph{et~al.}, ``{Estimation of Nakagami Distribution Parameters Based
  on Signal Samples Corrupted with Multiplicative and Additive Disturbances},''
  vol.~2, no. September, pp. 12--14, 2007.

\bibitem{Wang2012}
N.~Wang, S.~Member, X.~Song, S.~Member, and J.~Cheng, ``{Generalized Method of
  Moments Estimation of the Nakagami- m Fading Parameter},'' vol.~11, no.~9,
  pp. 3316--3325, 2012.

\bibitem{Zhang2002}
Q.~T. Zhang and S.~Member, ``{A Note on the Estimation of Nakagami-m Fading
  Parameter},'' vol.~6, no.~6, pp. 237--238, 2002.

\bibitem{DBLP:journals/corr/abs-1207-3510}
Q.~Wang, ``Hmrf-em-image: Implementation of the hidden markov random field
  model and its expectation-maximization algorithm,'' \emph{CoRR}, vol.
  abs/1207.3510, 2012.

\end{thebibliography}

 \end{document}